\documentclass{article}
\usepackage[preprint]{main}

\usepackage{microtype}
\usepackage{hyperref}
\usepackage{url}
\usepackage{booktabs}
\usepackage{graphicx}
\usepackage{amsmath}
\usepackage{placeins}
\usepackage{wrapfig}
\usepackage{float}
\usepackage{stfloats}
\usepackage[font=small,labelfont=bf]{caption}
\usepackage{tabularx}
\usepackage{lineno}

\definecolor{darkblue}{rgb}{0, 0, 0.5}
\hypersetup{colorlinks=true, citecolor=darkblue, linkcolor=darkblue, urlcolor=darkblue}

\title{Exploring the Limits of Pruning: Task-Specific Neurons, Model Collapse, and Recovery in Task-Specific Large\\Language Models}

\usepackage{authblk}

\author[1]{M.~K.~Khalidi~Siam}
\author[1]{Md.~Tausif-Ul-Islam}
\author[1]{Md.~Reshad~Romim~Khan}
\author[1]{Mohammed~Ali~Hossain}
\author[1]{Mushfiqul~Amin}
\author[1]{Labib~Hasan~Khan}
\author[1]{Niloy~Farhan}
\author[1]{Farig~Sadeque}

\affil[1]{BRAC University\par
mk.khalidi.siam@g.bracu.ac.bd}

\begin{document}
\ifcolmsubmission
\linenumbers
\fi
\maketitle
\begin{abstract}
Neuron pruning is widely used to reduce the computational cost and parameter footprint of large language models, yet it remains unclear whether neurons in task-specific models contribute uniformly to task performance. In this work, we provide empirical evidence for the existence and importance of task-specific neurons through a systematic pruning study on language models specialized for mathematical reasoning and code generation. We introduce an activation-based selectivity metric to identify neurons with low contribution to the target task and prune them while preserving target-task accuracy, and compare selective pruning with random pruning. Selective pruning consistently outperforms random pruning, indicating that activation-based selectivity provides a systematic advantage over random pruning. Reverse pruning experiments further show that removing a small subset of highly task-specific neurons ($\sim$10\%) causes complete performance collapse, suggesting that there exist task specific neuron and critical task information is concentrated in a small portion of the network. In contrast, selective pruning of less critical neurons ($\sim$30\% - $\sim$35\%) reduces accuracy but still preserves significant performance. We also observed consistent reductions in parameters and runtime VRAM usage, along with improved inference throughput as pruning increases. Experiments on both 1.5B and 7B models reveal a robustness threshold around 15–20\% pruning, beyond which accuracy loss and generation failures increase sharply. Fine-tuning substantially recovers performance across pruning levels, particularly for aggressively pruned models. These findings provide empirical evidence of neuron specialization in task-specific language models and offer insights into pruning robustness, model redundancy, and post-pruning recoverability.

\end{abstract}

\section{Introduction}
Large language models (LLMs) achieve strong performance across many NLP tasks, but their size and computational cost limit deployment in resource-constrained settings. Model compression techniques such as quantization, knowledge distillation, and pruning have gained attention, with neuron pruning being particularly promising as it removes redundant neurons while preserving performance.

Prior work, including SparseGPT~\citep{frantar2023sparsegpt}, SliceGPT~\citep{ashkboos2024slicegpt}, and LLM-Pruner~\citep{ma2023llmpruner}, focuses on pruning general-purpose models, with limited exploration of task-specific models. We investigate neuron pruning in task-specific models, studying whether performance can be preserved or recovered via post-pruning fine-tuning, and whether neuron importance exhibits task-specific characteristics.

Our approach ranks neurons based on activation for a target task while treating other tasks as distractors. We progressively prune the least active neurons across pruning ratios ($\sim$5\%–35\%) and analyze the impact on task accuracy, followed by post-pruning fine-tuning to evaluate recovery. Random pruning serves as a baseline, while reverse pruning removes the most active neurons to test the neuron's importance.

We evaluate 1.5B and 7B models to study scaling effects and analyze pruning-induced anomalies, including degeneration loops and unstable generation, to assess robustness.

Our work makes the following contributions:

\begin{itemize}

\item \textbf{Task-specific pruning analysis:} 
A systematic study of neuron pruning in task-specific language models across 1.5B and 7B scales.

\item \textbf{Performance recovery:} 
Evaluation of post-pruning fine-tuning for recovering task performance.

\item \textbf{Neuron importance via reverse pruning:}
Analysis of the impact of removing highly active (task-relevant) neurons.

\item \textbf{Abnormal behavior analysis:} 
Study of pruning-induced anomalies, including degeneration loops and unstable generation.

\end{itemize}

\section{Related Work}

Recent systems actively exploit the inherent sparsity of feed-forward network (FFN) activations; for example,~\citet{song2024powerinfer} show that neuron activations follow a power-law distribution, where some neurons are frequently active, while many others remain rarely activated.~\citet{wang2024sharing} reveal that while some neurons generalize across tasks, others show more task or context-specific behavior.

Because neurons exhibit task specialization, targeted neuron management has emerged as a powerful tool for precise capability control.~\citet{pochinkov2024dissecting} utilize selective neuron pruning for post-hoc machine unlearning, demonstrating that disabling specific neurons (by zeroing their parameters) degrades targeted skills while preserving the overall model architecture. 

To achieve tangible hardware acceleration, compression must be structural rather than purely theoretical. While earlier methods like SparseGPT~\citep{frantar2023sparsegpt} relied on unstructured weight masking, recent approaches focus on physical model reduction.~\citet{ashkboos2024slicegpt} introduce SliceGPT, which structurally compresses models by directly deleting rows and columns from weight matrices. Similarly, \citet{ma2023llmpruner} develope LLM-Pruner to remove coupled model structures based on gradient importance. We adopt the rigorous physical matrix slicing techniques seen in these works; however, instead of relying on complex gradient approximations, we employed activation based neuron pruning.

Aggressive structural pruning inevitably degrades model performance \citep{ma2023llmpruner}. To mitigate this degradation without the massive computational expense of continued pre-training \citep{xia2024sheared}, Parameter-Efficient Fine-Tuning (PEFT) has become the standard rehabilitation strategy. Frameworks such as LoRAPrune \citep{zhang2024loraprune} successfully employ Low-Rank Adaptation (LoRA) to restore lost capabilities. Aligning with this prior work, we integrate LoRA fine-tuning to rehabilitate the generative stability and task-specific reasoning pathways disrupted during our activation-based pruning.

Prior studies \citep{dalvi-etal-2020-analyzing} have primarily focused on GELU-based transformer models, such as BERT and XLNet, where a large fraction of neurons (around 85\%) were found to be functionally redundant and at least 92\% of them can be removed when optimizing models for downstream tasks. Recent theoretical frameworks establish that high parameter redundancy is a fundamental property of network scaling rather than an empirical artifact. Corresponding “redundancy laws” imply that task adaptation requires substantially less model capacity~\citep{bi2025scalinglawsredundancy}.

\section{Methodology}

\subsection{Model Selection and Dataset Construction}
\label{sec:3.1}

We evaluated our approach across multiple task by selecting models for both mathematical reasoning and code generation. For mathematical reasoning, we used Qwen2.5-Math-7B-Instruct and Qwen2.5-Math-1.5B-Instruct~\citep{yang2024qwen25mathtechnicalreportmathematical}. For code generation, we selected Qwen2.5-Coder-7B-Instruct and Qwen2.5-Coder-1.5B-Instruct~\citep{hui2024qwen25codertechnicalreport}. We included both smaller and larger models to assess whether the performance differences introduced by our approach remained consistent across parameter scales.

Datasets were categorized according to task relevance, as summarized in Table~\ref{tab:datasets}. For mathematical reasoning, GSM8K~\citep{cobbe2021gsm8k} served as the primary target dataset. For code generation, we constructed a target dataset from CodeFeedback-Filtered-Instruction~\citep{codefeedback_filtered_instruction_dataset}, restricting samples to Python programs with a maximum token length of 1k.

Datasets representing passage-based question answering and dialogue tasks specifically SQuAD~\citep{rajpurkar-etal-2016-squad} and everyday-Conversational-cleaned~\citep{everydayConversationalCleaned2024} were treated as distractors to represent non-target semantics.

We intentionally did not treat math data as distractors for code models, nor code data as distractor for math models, since prior work suggests mathematical reasoning and code generation share underlying structures and may transfer positively across task rather than act as purely irrelevant signal~\citep{shao2024deepseekmathpushinglimitsmathematical}.

\begin{table}[htbp]
\centering
\scriptsize
\setlength{\tabcolsep}{3pt}

\begin{tabularx}{\columnwidth}{@{}l c l c X@{}}
\toprule
\textbf{Dataset} & \textbf{Train set(capture activation)} & \textbf{Task Type} & \textbf{Fine-tune set} & \textbf{Benchmark set} \\
\midrule
GSM8K & 800 prompts & Math & Official Train set (7.5k) & Official Test set (1.3k) \\
CodeFeedback-Filtered & 800 prompts & Code & 7k & HumanEval (pass@1) \\
SQuAD & 800 prompts & QnA & - & 200 prompts \\
everyday-Conversational-cleaned & 800 prompts & Dialogue & - & 200 prompts \\
\bottomrule
\end{tabularx}

\caption{Overview of Datasets}
\label{tab:datasets}
\end{table}

Note that, training sets were used exclusively to capture neuron activations, while benchmark sets were reserved for benchmarking. For the math dataset (GSM8K), the training set is the subset of official train set.

\subsection{Activation Capture for Target and Distractor}
Prompts from the training sets were passed through the model to extract activations (only considered input tokens) from each feed-forward network (FFN) layer. For each transformer layer $\ell$, we focused on the intermediate SwiGLU activation $h^{(\ell)}$ before the down-projection, as it captured the critical interaction between the gate and up-projected input:
{\small
\begin{equation}
    h^{(\ell)} = \sigma \!\left( W^{(\ell)}_{\text{gate}} x^{(\ell)} \right) \odot \left( W^{(\ell)}_{\text{up}} x^{(\ell)} \right).
\end{equation}
}

To reduce memory overhead and processing demands, we computed the element-wise absolute value of the MLP intermediate activations and then calculated the mean only over valid (non-padded) tokens in each sequence. This produced a single, prompt-level mean activation vector for every layer, ensuring that padding tokens do not influence the neuron statistics.

\subsection{Neuron Selectivity}
For each neuron, a selectivity score was calculated as the standardized difference in average activations between target and distractor prompts. Let $\mu^{(\ell)}_{\text{target}, n}$ and $\mu^{(\ell)}_{\text{distractor}, n}$ represent the mean activations of neuron $n$ in layer $\ell$ across all target and distractor sequences, respectively. The directional selectivity score is defined as:

{\small
\begin{equation}
    S^{(\ell)}_n = \frac{\mu^{(\ell)}_{\text{target}, n} - \mu^{(\ell)}_{\text{distractor}, n}}{\sigma^{(\ell)}_n + \epsilon},
    \label{eq:selectivity}
\end{equation}
}

where $\sigma^{(\ell)}_n$ is the per-neuron standard deviation over all prompts (pooling target and distractor) and $\epsilon = 10^{-6}$ ensures numerical stability. Positive scores indicate target relevance, while negative scores signal distractor association, making neurons pruning candidates.

\subsection{Structural Pruning}
We applied structural pruning by physically removing neurons from the network, targeting only the MLP blocks. In Transformer architectures, the feed-forward dimension typically satisfied $d_{ff} \approx 4 d_{model}$, making it the dominant contributor to model parameters. In prior work, feed-forward layers exhibited substantial redundancy and could be simplified with minimal performance degradation, whereas modifying attention layer has larger impact on performance~\citep{pires-etal-2023-one}. Therefore, we did not modify the attention layer. To preserve the original uniform hidden dimension across layers, we pruned neurons uniformly across all MLP blocks. Our goal is to prune distractor-friendly neurons (i.e., those with negative selectivity); however, since pruning is performed by removing the bottom 5\%–35\% of neurons, some neurons with positive selectivity may also be removed. Additionally, as we apply uniform pruning, the threshold $\theta^{(\ell)}$ is identical across all layers.

To maximize hardware utilization, structural pruning was applied uniformly across all MLP layers such that the remaining neuron count remained strictly divisible by 128, optimizing matrix multiplication efficiency within GPU kernels~\citep{nvidia2023matmult}. Consequently, exact pruning ratios deviated slightly (Table~\ref{tab:pruning_ratios}); for simplicity, we reported them using nominal 5\% increments.

\begin{wraptable}[7]{r}{0.4\textwidth}
\vspace{-10pt}
\centering
\resizebox{\linewidth}{!}{
\begin{tabular}{@{}ccc@{}}
    \toprule
    \textbf{Nominal Ratio} & \textbf{Qwen-7B Exact} & \textbf{Qwen-1.5B Exact} \\
    \midrule
    5\%  & 5.41\%  & 5.71\%  \\
    10\% & 10.14\% & 10.00\% \\
    15\% & 15.54\% & 15.71\% \\
    20\% & 20.27\% & 20.00\% \\
    25\% & 25.00\% & 25.71\% \\
    30\% & 30.41\% & 30.00\% \\
    35\% & 35.14\% & 35.71\% \\
    \bottomrule
\end{tabular}
}
\captionsetup{font=scriptsize}
\caption{Exact Pruning Ratios to Maintain 128-Divisibility}
\label{tab:pruning_ratios}
\vspace{-10pt}
\end{wraptable}
Specifically, neurons satisfying $S^{(\ell)}_n \le \theta^{(\ell)}$ formed the pruning set $\mathcal{P}^{(\ell)}$, while the remainder formed the survivor set $\mathcal{K}^{(\ell)}$. Pruning physically removed the neurons in $\mathcal{P}^{(\ell)}$ by slicing the corresponding rows from the gate and up-projection weight matrices ($W^{(\ell)}_{\text{gate}}$, $W^{(\ell)}_{\text{up}}$), and the corresponding columns from the down-projection matrix ($W^{(\ell)}_{\text{down}}$). This uniformly reduced the intermediate dimension of the MLP from its original size to the size of the survivor set $\lvert \mathcal{K}^{(\ell)} \rvert$.

\subsection[Section 3.5]{LoRA Fine-Tuning Integration}
\label{sec:3.5}
To recover task-specific performance after pruning, we used Low-Rank Adaptation (LoRA)~\citep{hu2021loralowrankadaptationlarge}, applying adapters to the projection layers of both self-attention and feed-forward blocks. Code models were fine-tuned with $r=16$ and a learning rate of $2 \times 10^{-4}$, while math models used $r=8$ and $2 \times 10^{-5}$. All models trained for 2 epochs with Bfloat16 precision.

\subsection{Evaluation Metrics}
For the Qwen-Math models, performance was evaluated using exact match (EM) accuracy on LaTeX expressions, extracted via a regex targeting \verb|\boxed{...}|. For the Qwen-Code models, functional correctness was assessed on HumanEval~\citep{chen2021evaluatinglargelanguagemodels} using $\text{pass@}1$, i.e., the fraction of generated functions that pass hidden unit tests on the first attempt. Due to resource constraints, we conducted a single run and report only pass@1.

Distractor tasks for both models included QnA and dialogue, evaluated via BERTScore~\citep{zhang2020bertscoreevaluatingtextgeneration} and SBERT cosine similarity~\citep{reimers-gurevych-2019-sentence}, respectively. Averaging these metrics yielded a unified distractor semantic similarity score (SSS).

In addition to task-specific performance, we evaluated structural instability induced by pruning by monitoring the frequency of generative failures. Specifically, we tracked anomalous behaviors such as missing end-of-sequence (EOS) tokens and repetitive degeneration loops, collectively referred to as "traps".

Note that, deterministic greedy generation (temperature $T = 0$) was used to ensure reproducibility and to accurately capture the effects of pruning without stochastic interference. 

Detailed formulations of neuron activation and pruning procedures, full LoRA hyperparameters, as well as hardware setup and evaluation metric configurations, are provided in the supplementary material.

\section{Result Analysis}

This section analyzes the behavior of models across different approach(i.e selective pruning, random pruning, reverse pruning), model's layer wise neuron distribution, trap analysis and performance recovery post pruning LoRA finetuning. We evaluate both mathematical reasoning and code
generation models using Exact Match (EM) and HumanEval scores,
respectively. We also report resource usage and inference throughput to evaluate the computational impact of different pruning levels.


\subsection{Pruning Analysis}

We prune the model from 5\% to 35\% to examine performance degradation and behavioral changes across pruning levels. While performance declines progressively with increased pruning, the degradation remains relatively mild at lower pruning ratios. However, beyond approximately 15\%--20\% pruning, the decline becomes significantly steeper, indicating a critical threshold in model robustness.

\begin{table}[H]
\centering
\scriptsize
\renewcommand{\arraystretch}{0.9}

\begin{minipage}{0.48\linewidth}
\centering
\textbf{(a) Performance Degradation Under Selective Pruning}

\vspace{3pt}

\resizebox{\linewidth}{!}{
\begin{tabular}{lcccc}
\toprule
Pruning & Qwen-7B Math & Qwen-1.5B Math & Qwen-7B Code & Qwen-1.5B Code \\
\midrule
Original & 94.47 & 84.0 & 85.37 & 62.80 \\
5\%  & 93.25 & 82.49 & 81.71 & 47.56 \\
10\% & 92.19 & 82.11 & 79.27 & 45.12 \\
15\% & 89.16 & 76.5 & 74.39 & 38.41 \\
20\% & 85.67 & 68.92 & 65.85 & 32.93 \\
25\% & 82.87 & 63.0 & 59.76 & 21.95 \\
30\% & 68.69 & 47.38 & 51.22 & 19.51 \\
35\% & 57.85 & 22.37 & 43.29 & 6.71 \\
\bottomrule
\end{tabular}
}
\end{minipage}
\hfill
\begin{minipage}{0.48\linewidth}
\centering
\textbf{(b) Trap Percentage Across Pruning Levels}

\vspace{3pt}

\resizebox{\linewidth}{!}{
\begin{tabular}{lcccc}
\toprule
Pruning & Qwen-7B Math & Qwen-1.5B Math & Qwen-7B Code & Qwen-1.5B Code \\
\midrule
Original & 0.45 & 0.91 & 0.0 & 0.0 \\
5\%  & 0.68 & 1.21 & 0.0 & 0.0 \\
10\% & 2.43 & 1.44 & 2.0 & 2.5 \\
15\% & 16.91 & 4.4 & 1.5 & 7.5 \\
20\% & 27.52 & 5.61 & 6.0 & 18.0 \\
25\% & 45.72 & 10.61 & 7.5 & 52.5 \\
30\% & 99.7 & 28.05 & 11.5 & 54.0 \\
35\% & 99.55 & 58.53 & 13.0 & 92.0 \\
\bottomrule
\end{tabular}
}
\end{minipage}

\caption{Pruning analysis results. EM for math models and HumanEval for code models.}
\label{tab:pruning_results}
\end{table}

Performance loss is measured using Exact Match (EM) for mathematical reasoning and human evaluation scores for coding tasks. In parallel, we also assess instability through trap-based evaluation. Notably, trap occurrences increase sharply beyond the 15\%--20\% pruning range, coinciding with the steeper performance drop. This suggests that moderate pruning not only reduces task accuracy but also disrupts the model’s ability to terminate generation properly, revealing a structural stability threshold around this pruning level.

In some cases, trap rates remain high even when accuracy is high, as the model first produces the correct answer and subsequently enters a trap state (Type 1; see ~\nameref{sec:trap-types} for details). We therefore count such instances as correct for accuracy while also recording them as trap occurrences.
\FloatBarrier
\subsection{Fine-Tuning Analysis}

After pruning, model performance declines as expected. We then investigate how much of this loss can be recovered through fine-tuning. Promisingly, both small and large models across math and code tasks show clear performance improvements after fine-tuning.The overall pattern of recovery reflects the earlier degradation trend, suggesting that models retain an inherent capacity to regain lost capabilities through fine-tuning. 

\begin{table*}[htbp]
\centering
\scriptsize
\renewcommand{\arraystretch}{0.9}

\begin{minipage}{0.48\textwidth}
\centering
\textbf{(a) Performance After Fine-Tuning}

\vspace{3pt}

\resizebox{\linewidth}{!}{
\begin{tabular}{lcccc}
\toprule
Fine-Tune & Qwen-7B Math & Qwen-1.5B Math & Qwen-7B Code & Qwen-1.5B Code \\
\midrule
Original & 94.47 & 84.0 & 85.37 & 62.80 \\
5\%  & 93.1 & 82.64 & 82.93 & 48.78 \\
10\% & 92.57 & 82.18 & 82.32 & 50.61 \\
15\% & 90.3 & 78.7 & 82.93 & 51.22 \\
20\% & 88.25 & 74.37 & 78.05 & 44.51 \\
25\% & 86.05 & 69.52 & 78.05 & 43.29 \\
30\% & 80.59 & 63.46 & 71.95 & 32.32 \\
35\% & 73.84 & 51.25 & 68.29 & 31.10 \\
\bottomrule
\end{tabular}
}
\end{minipage}
\hfill
\begin{minipage}{0.48\textwidth}
\centering
\textbf{(b) Trap Percentage After Fine-Tuning}

\vspace{3pt}

\resizebox{\linewidth}{!}{
\begin{tabular}{lcccc}
\toprule
Fine-Tune & Qwen-7B Math & Qwen-1.5B Math & Qwen-7B Code & Qwen-1.5B Code \\
\midrule
Original & 0.45 & 0.91 & 0.0 & 0.0 \\
5\%  & 0.83 & 1.14 & 0.0 & 0.0 \\
10\% & 2.5 & 1.21 & 0.5 & 0.5 \\
15\% & 10.39 & 1.14 & 0.0 & 0.5 \\
20\% & 5.38 & 2.27 & 0.0 & 0.5 \\
25\% & 5.91 & 3.49 & 0.0 & 0.5 \\
30\% & 22.82 & 4.17 & 0.5 & 1.0 \\
35\% & 18.12 & 10.54 & 0.0 & 1.5 \\
\bottomrule
\end{tabular}
}
\end{minipage}

\caption{Fine-tuning analysis results. EM for math models and HumanEval for code models.}
\end{table*}

In general, recovery follows an upward trend: models with lower pruning ratios often achieve better performance after fine-tuning. However, this relationship is not strictly monotonic. For example, in the Qwen 1.5B code model, the 10\% and 15\% pruned variants outperform the 5\% pruned model after fine-tuning. This demonstrates that lower pruning levels do not always guarantee superior final accuracy, and that moderately pruned models can sometimes surpass lightly pruned ones after fine-tuning.

Consistent with our pruning analysis, we evaluate both accuracy and trap count during fine-tuning. Fine-tuning substantially improves accuracy while reducing trap occurrences, indicating that it not only restores task performance but also reduces post-pruning induced instability, making it a promising approach for mitigating pruning-related traps.
\FloatBarrier
\subsection{Impact of Pruning Severity on Smaller vs. Larger Models}

We analyze pruning resilience of small and large models,
focusing on Qwen-1.5B and Qwen-7B. Larger models
consistently tolerate higher neuron removal with smaller
performance degradation, particularly under aggressive pruning.
Relative accuracy loss ensures fair comparison across
differing baselines:
{\small
\begin{equation}
\text{Relative Accuracy Loss (\%)} = \frac{A_{\text{original}} - A_{\text{pruned}}}{A_{\text{original}}} \times 100
\end{equation}
}
\begin{wrapfigure}[10]{r}{0.5\textwidth}
\centering
\includegraphics[width=0.5\textwidth]{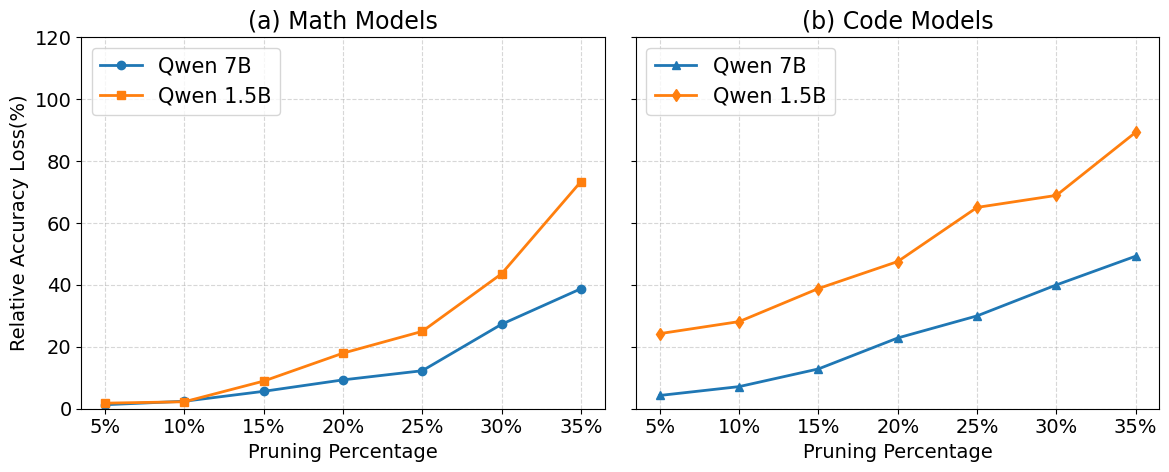}
\captionsetup{font=scriptsize}
\caption{Relative accuracy loss across pruning levels for smaller and larger models.}
\label{fig:scale}
\end{wrapfigure}
where $A_{\text{original}}$ represents the performance of the
model before pruning and $A_{\text{pruned}}$ denotes the
performance immediately after pruning.

Even at 35\% pruning, larger models maintain stability
despite more neurons being removed in absolute terms.
This reflects higher representational capacity and redundancy.
Smaller models, with fewer redundant neurons, lose critical capacity more quickly as pruning increases, leading to larger performance loss.


\subsection{Impact of Pruning Severity on Fine-Tuning Gains}

We study how pruning severity affects performance recovery after fine-tuning.
Relative performance gain quantifies how much capability is restored through
LoRA adaptation following structural pruning:

{\small
\begin{equation}
\text{Relative Gain (\%)} =
\frac{A_{\text{after}} - A_{\text{before}}}{A_{\text{before}}} \times 100
\end{equation}
}

\begin{wrapfigure}[7]{r}{0.45\textwidth}
\vspace{-9pt}
\centering
\includegraphics[width=0.5\textwidth]{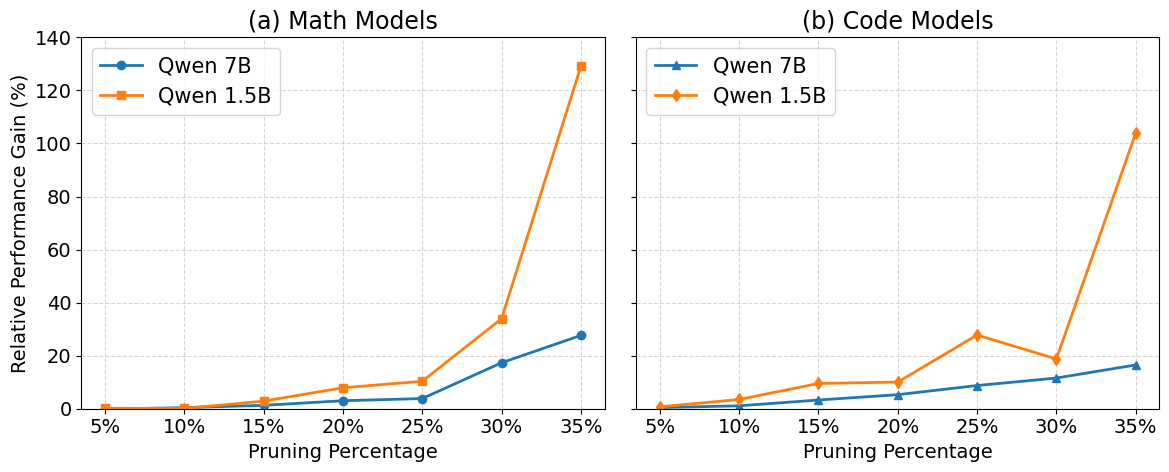}
\captionsetup{font=scriptsize}
\vspace{-20pt}
\caption{Relative gain after fine-tuning across pruning levels; code model values scaled down ×3.5 for visualization clarity.}
\label{fig:gain}
\end{wrapfigure}
where $A_{\text{before}}$ represents performance immediately after
pruning and $A_{\text{after}}$ denotes performance after fine-tuning.

Lightly pruned models yield minimal gains, while heavily pruned models show substantial performance improvement. For example, 35\% pruning leads to the largest relative gains after fine-tuning, particularly in smaller models.

\subsection{Selective Pruning vs. Random Pruning}

Selective pruning removes neurons with low relevance to the target task while preserving those most important for performance, enabling compression with minimal accuracy loss. A possible counterargument is that neurons may be functionally neutral, in which case selective pruning would perform similarly to random pruning. To test this, we compare selective pruning with random pruning, defining the performance difference as:

{\small
\begin{equation}
\Delta A = A_{\text{selective}} - A_{\text{random}}
\end{equation}
}

\begin{wrapfigure}[13]{r}{0.48\columnwidth}
\vspace{-8pt}
\centering

\includegraphics[width=\linewidth]{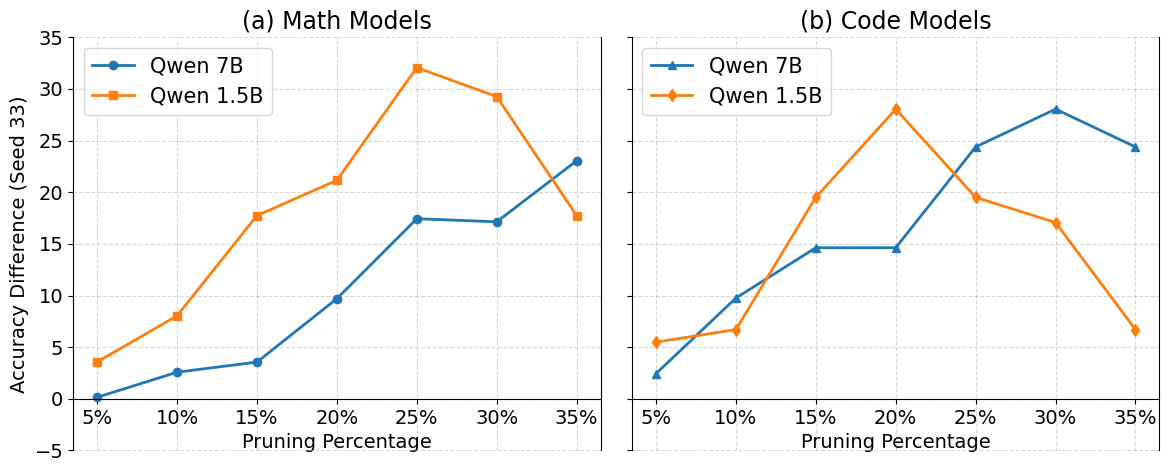}

\vspace{6pt}

\includegraphics[width=\linewidth]{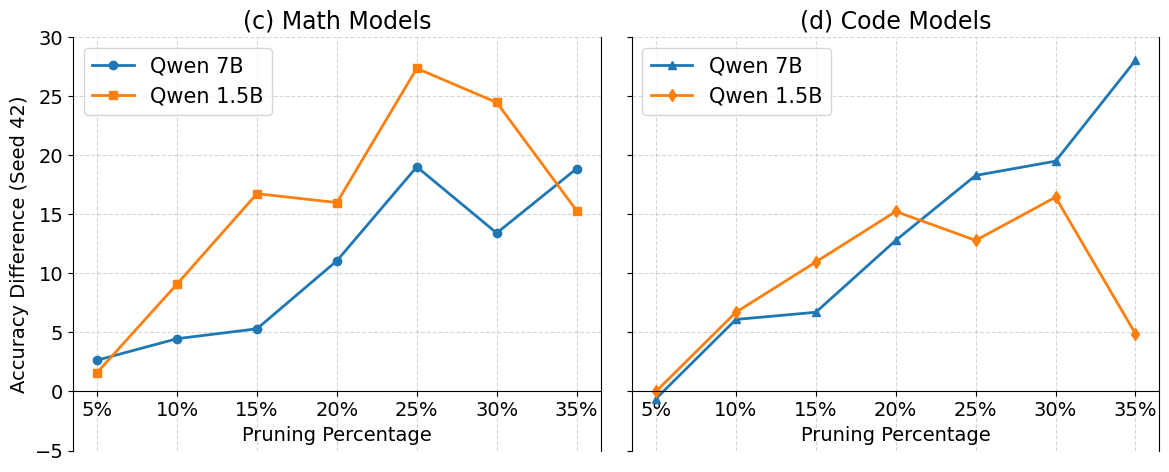}

\captionsetup{font=scriptsize}
\caption{Selective vs.\ random pruning under different seeds.}
\label{fig:random}
\vspace{-8pt}
\end{wrapfigure}

where $A_{\text{selective}}$ and $A_{\text{random}}$ denote performance after selective and random pruning, respectively.

Random pruning is evaluated with two independent seeds (42 and 33) to reduce stochastic bias and improve reproducibility. Across all models and pruning levels, selective pruning consistently retains higher performance, while random pruning leads to greater degradation. These results indicate that neuron importance is non-uniform and that activation-based selection provides a clear advantage over random pruning.

\subsection{Reverse Pruning}

To further examine neuron importance, we introduce reverse
pruning, where neurons with the highest target-task selectivity are
intentionally removed. This experiment serves as a counterfactual
test to determine how strongly model performance depends on
task-relevant neurons.

\begin{wrapfigure}[10]{r}{0.49\textwidth}
\vspace{-8pt} 
\centering
\includegraphics[width=0.50\textwidth]{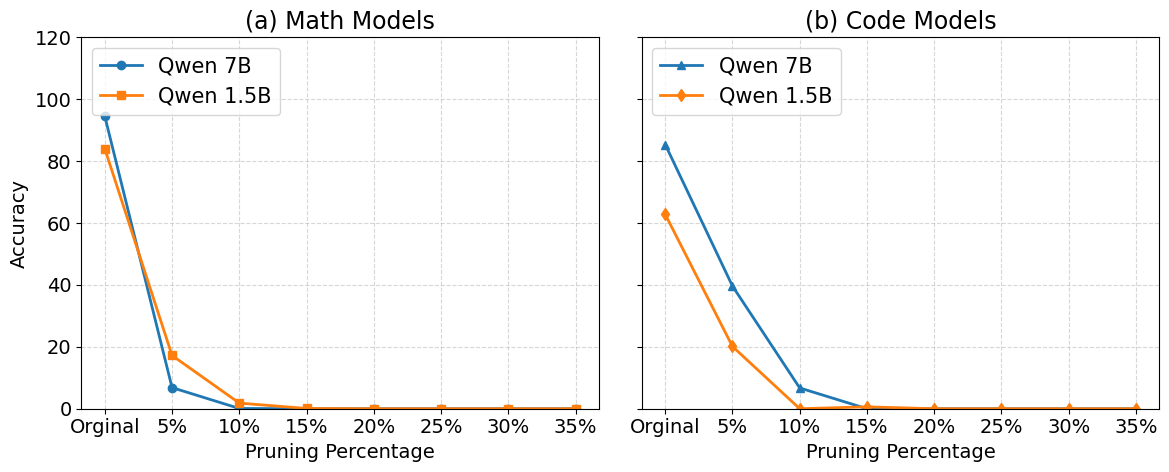}
\captionsetup{font=scriptsize}
\caption{Comparison between selective and reverse pruning strategies.}
\label{fig:reverse}
\end{wrapfigure}

Unlike selective pruning, which removes distractor-associated neurons, reverse pruning deliberately removes task-specific neurons. Results show rapid performance degradation: models approach critical failure at approximately 5\% of reverse pruning and collapse entirely near 10\%. In this context, collapse refers to a complete loss of task accuracy, accompanied by frequent failure modes such as falling into reasoning traps and generating responses that are meaningless, incoherent, or nonsensical. In contrast, selective pruning produces gradual accuracy decline without complete breakdown even at 35\% pruning, as summarized in Table~\ref{tab:pruning_results}

These findings indicate that model capability depends on a relatively
small subset of highly task-relevant neurons, while the remaining
neurons provide supportive or redundant representations. Together
with the selective versus random pruning results, this experiment
provides strong evidence for the existence of specialized
task-specific neurons in task-specific language models.

\subsection[Section 4.7]{Traps and Structural Instability in Pruned Models}
\label{sec:trap-types}
To evaluate natural structural stability, we performed unconstrained generation up to 1024 tokens. Original models rarely exhibited traps ($<1\%$), while trap frequency increased with pruning ratio. We observe two types of traps: 

\begin{itemize}
    \item \textbf{Type 1: Degeneration Loop with Complete Reasoning:} The model produces meaningful reasoning and an answer with proper formatting but then fails to emit an EOS token, continuing with repeated symbols, tokens, or sequences.
    
    \item \textbf{Type 2: Degeneration Loop with Incomplete Reasoning:} The model fails to perform meaningful reasoning, collapsing early or mid-generation into incoherent and nonsensical text such as repeated symbols, tokens, or sequences. As the pruning ratio increases, the frequency of Type 2 traps also increases.
\end{itemize}

Qwen2.5-Math-7B-Instruct exhibits more traps but also higher accuracy, as most are Type-1. In practice, this can be mitigated with custom stopping criteria. We report unconstrained results to highlight these architectural dynamics, and provide one example per trap type, plus a mitigated Type-1 sample in the \nameref{sec:trap_examples} section.

Overall, pruning disrupts EOS prediction and reasoning stability, introducing generation loops largely absent in the original models.

\subsection{Distractor Analysis}

\setlength{\intextsep}{4pt}

\begin{wrapfigure}[14]{r}{0.5\columnwidth} 
\vspace{-12pt} 
\centering
\includegraphics[width=\linewidth]{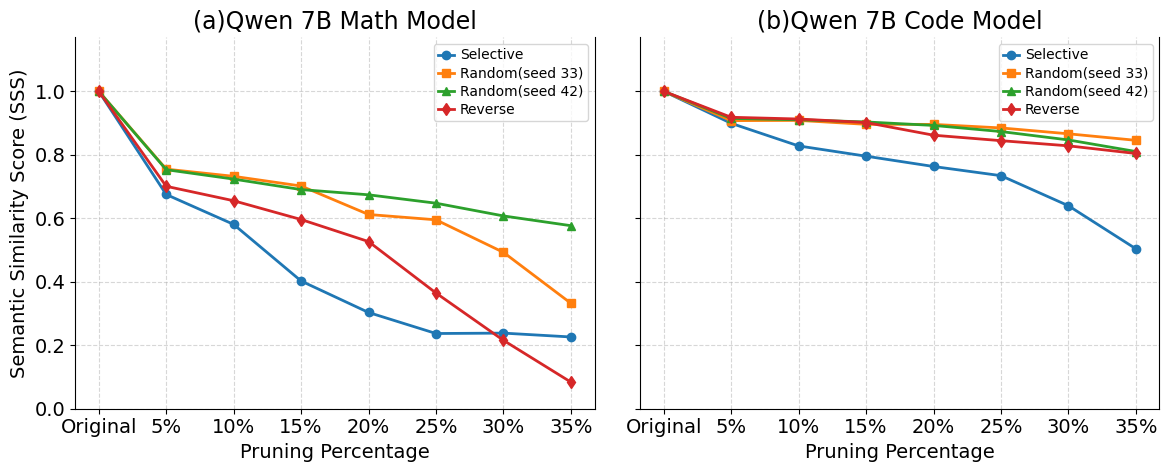}
\vspace{-2pt} 
\includegraphics[width=\linewidth]{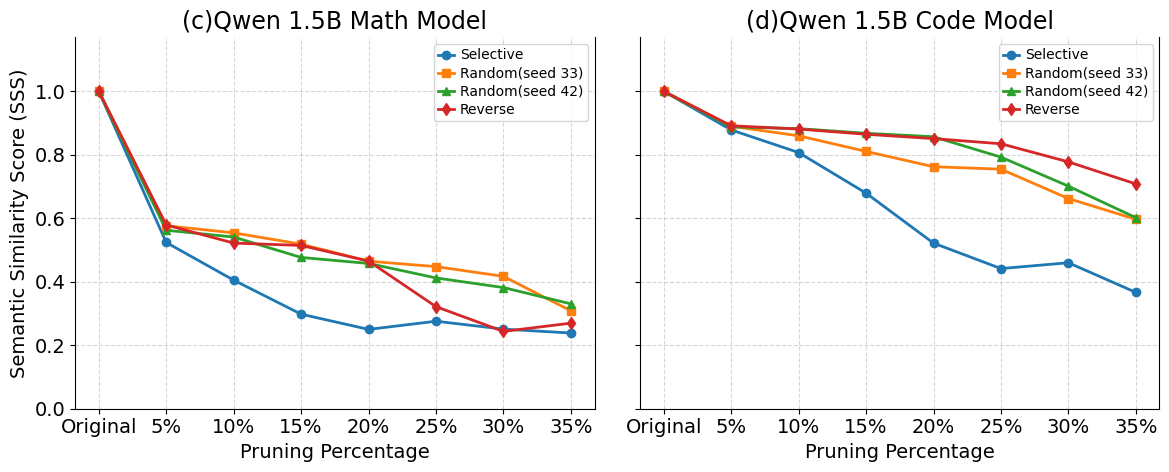}
\vspace{-6pt} 
\captionsetup{font=scriptsize}
\vspace{-6pt} 
\caption{Distractor semantic similarity analysis}
\label{fig:semantic_similarity}
\end{wrapfigure}

To further examine task-specific neurons, we compare random, reverse, and selective pruning on the distractor tasks. Previously, reverse pruning causes a sharp performance drop on the target task, suggesting that a small subset of neurons is strongly target-relevant. In contrast, distractor performance should degrade less under reverse pruning since distractor-friendly neurons are largely preserved.

We measure pruning-induced degradation using semantic similarity between the pruned and original model outputs (baseline), defined as:
\vspace{16pt} 
{\small
\begin{equation}
\text{Semantic Similarity} = \frac{\text{BERTScore} + \text{SBERT Cosine Similarity}}{2}
\end{equation}
}
where BERTScore is used for QnA and SBERT cosine similarity for dialogue.

As shown in Figure~\ref{fig:semantic_similarity}, selective pruning consistently leads to the largest degradation. Notably, reverse pruning reduces target-task performance to nearly zero after only $10\%$ pruning as show in Figuer~\ref{fig:reverse}, while distractor performance remains relatively stable. These results support the existence of task-specific neurons and show that selective pruning can effectively identify and prune neurons based on activation.

\subsection{Layer-wise Neuron Distribution Analysis}
\setlength{\intextsep}{4pt}

\begin{wrapfigure}[11]{r}{0.43\columnwidth} 
\vspace{-8pt} 
\centering
\includegraphics[width=\linewidth]{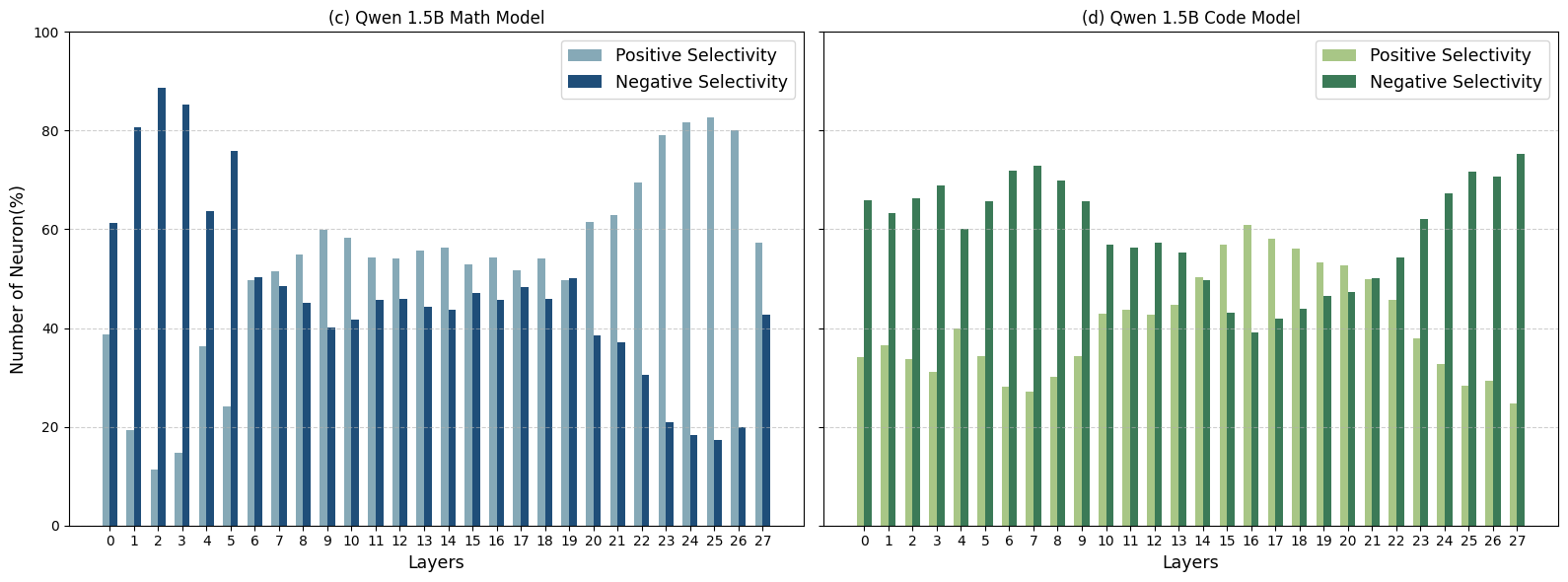}
\vspace{-2pt} 
\includegraphics[width=\linewidth]{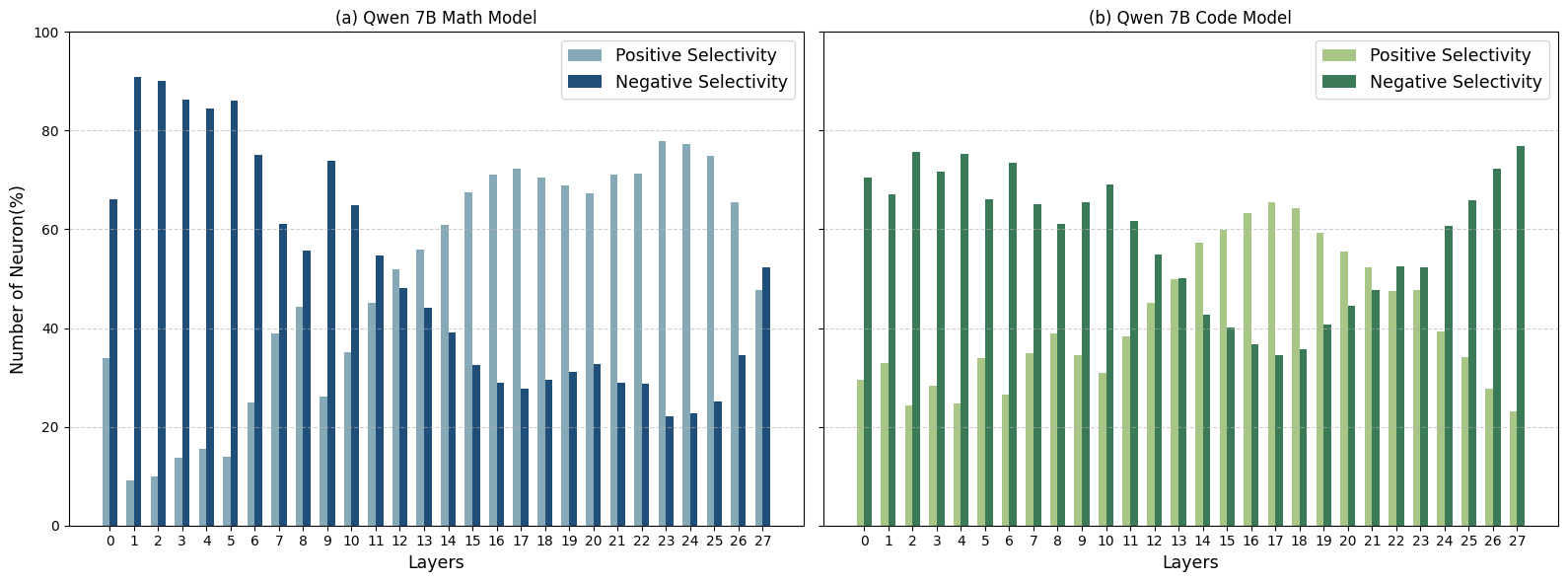}
\vspace{-20pt} 
\captionsetup{font=scriptsize}
\caption{Layer-wise neuron distribution (Target vs Distractor)}
\label{fig:layer}
\vspace{-5pt}
\end{wrapfigure}

Neuron activation is non-uniform across layers for both math and code models. In both cases, early layers are predominantly distractor-oriented, while middle layers shift toward target-relevant activations, indicating a transition from general to task-specific processing. 

In later layers, math models remain strongly target-dominant, whereas code models exhibit a more mixed behavior with a tendency to shift back toward distractor dominance. This suggests that task-specific specialization is more consistently preserved in math models than in code models.

Despite this non-uniform distribution, we adopt uniform pruning across layers to preserve the original model structure, as the feed-forward dimensions are consistent across layers.

\subsection{Resource and Throughput Analysis}

Physical pruning reduces the number of model parameters, leading to a smaller model size and fewer FLOPs. As a result, runtime VRAM usage decreases while inference throughput (tokens/s) increases as pruning level increases. 

\begin{wraptable}[7]{r}{0.4\textwidth}
\vspace{-9pt}
\centering
\footnotesize
\renewcommand{\arraystretch}{1.05}

\resizebox{\linewidth}{!}{
\begin{tabular}{c cc rr}
\hline
& \multicolumn{2}{c}{VRAM Usage (GB)} & \multicolumn{2}{c}{Throughput (Token/s)} \\
\cline{2-5}
Pruning (\%) & Qwen-7B & Qwen-1.5B & Qwen-7B & Qwen-1.5B \\
\hline
0  & 14.26 & 2.90 & 35.83 & 46.21 \\
5  & 13.65 & 2.78 & 37.52 & 46.50 \\
10 & 13.15 & 2.71 & 38.50 & 47.31 \\
15 & 12.64 & 2.56 & 39.27 & 48.15 \\
20 & 12.16 & 2.55 & 39.94 & 48.81 \\
25 & 11.69 & 2.39 & 40.34 & 49.01 \\
30 & 11.06 & 2.26 & 40.92 & 49.39 \\
35 & 10.56 & 2.14 & 42.70 & 50.07 \\
\hline
\end{tabular}
}

\captionsetup{font=scriptsize}
\caption{Resource Usage and Throughput Across Pruning Levels (batch size = 1)}
\vspace{-10pt}
\end{wraptable}

However, aggressive pruning, although improving resource efficiency, can negatively affect model accuracy and stability, indicating a trade-off between resource reduction and model performance. Note that tokens/s may vary depending on the hardware and background processes running on the OS.

\section{Discussion and Limitations}

\subsection{Pruning and Reduced Accuracy}

Our selectivity metric isolates task-specific neurons by measuring the relative activation difference between target ($a^T_i$) and distractor ($a^D_i$) contexts, rather than absolute neuron utility, as defined in equation~\ref{eq:selectivity}. Consequently, neurons that contribute moderately to both domains may receive low selectivity scores and be removed during structural pruning.

For example, neurons with $(a^T_i = 0.5, a^D_i = 0.8)$ and $(a^T_i = 0.1, a^D_i = 0.4)$ have the same relative difference and may get identical selectivity scores, even though the first has higher target activation. Similarly, $(a^T_i = 0.6, a^D_i = 0.9)$ may score lower than $(a^T_i = 0.2, a^D_i = 0.4)$ despite stronger target activation.

As pruning severity increases, the removal of these moderately useful neurons accumulates and gradually disrupts distributed representations. Early pruning removes redundancy with little impact, but aggressive pruning eliminates neurons that still contribute to the task, causing nonlinear degradation and reduced downstream performance.

\subsection{Fine-Tuning Scope and Dataset Diversity}
The fine-tuning protocol here serves as a proof of concept for capability recovery, not an exhaustive search for optimal performance. Prior work shows that diverse datasets improve generalization by promoting robust, transferable representations, while fine-tuning on small datasets (e.g., only the GSM8K training split) limits broader generalization~\citep{yu-etal-2022-data, li2025macromicroprobingdataset}. In our case, the limited diversity of fine-tuning data likely constrained post-pruning recovery, and a more diverse corpus could enable stronger, more stable generalization.

\section{Conclusion}
Our experiments show that task-specific language models exhibit neuron specialization, where only a subset of neurons contributes strongly during inference ($\sim$10\%). Reverse pruning of these critical neurons causes rapid performance degradation and complete collapse after removing only $\sim$10\%. In contrast, selectively pruning less important neurons leads to gradual decline, with the model retaining substantial capability even at higher pruning levels.

Distractor analysis further confirms neuron specialization: reverse pruning primarily removes target-specific neurons, so performance on distractor tasks remains largely stable, while selective pruning of distractor-friendly neurons causes larger drops. These results support the existence of task-specific neurons and show that selective pruning can effectively identify and prune neurons based on activation. We also observe that aggressive pruning introduces inference instability, increasing generation traps. LoRA-based fine-tuning effectively recovers performance and mitigates these failures, highlighting its utility for post-pruning recovery in task-specific models.

\bibliography{main}

\begin{thebibliography}{25}
\providecommand{\natexlab}[1]{#1}
\providecommand{\url}[1]{\texttt{#1}}
\expandafter\ifx\csname urlstyle\endcsname\relax
  \providecommand{\doi}[1]{doi: #1}\else
  \providecommand{\doi}{doi: \begingroup \urlstyle{rm}\Url}\fi

\bibitem[Ashkboos et~al.(2024)]{ashkboos2024slicegpt}
Saleh Ashkboos et~al.
\newblock Slicegpt: Compress large language models by deleting rows and columns.
\newblock In \emph{The Twelfth International Conference on Learning Representations (ICLR)}, 2024.

\bibitem[Bi \& Calhoun(2025)Bi and Calhoun]{bi2025scalinglawsredundancy}
Yuda Bi and Vince~D Calhoun.
\newblock Scaling laws are redundancy laws, 2025.
\newblock URL \url{https://arxiv.org/abs/2509.20721}.

\bibitem[Chen et~al.(2021)Chen, Tworek, Jun, Yuan, de~Oliveira~Pinto, Kaplan, Edwards, Burda, Joseph, Brockman, Ray, Puri, Krueger, Petrov, Khlaaf, Sastry, Mishkin, Chan, Gray, Ryder, Pavlov, Power, Kaiser, Bavarian, Winter, Tillet, Such, Cummings, Plappert, Chantzis, Barnes, Herbert-Voss, Guss, Nichol, Paino, Tezak, Tang, Babuschkin, Balaji, Jain, Saunders, Hesse, Carr, Leike, Achiam, Misra, Morikawa, Radford, Knight, Brundage, Murati, Mayer, Welinder, McGrew, Amodei, McCandlish, Sutskever, and Zaremba]{chen2021evaluatinglargelanguagemodels}
Mark Chen, Jerry Tworek, Heewoo Jun, Qiming Yuan, Henrique~Ponde de~Oliveira~Pinto, Jared Kaplan, Harri Edwards, Yuri Burda, Nicholas Joseph, Greg Brockman, Alex Ray, Raul Puri, Gretchen Krueger, Michael Petrov, Heidy Khlaaf, Girish Sastry, Pamela Mishkin, Brooke Chan, Scott Gray, Nick Ryder, Mikhail Pavlov, Alethea Power, Lukasz Kaiser, Mohammad Bavarian, Clemens Winter, Philippe Tillet, Felipe~Petroski Such, Dave Cummings, Matthias Plappert, Fotios Chantzis, Elizabeth Barnes, Ariel Herbert-Voss, William~Hebgen Guss, Alex Nichol, Alex Paino, Nikolas Tezak, Jie Tang, Igor Babuschkin, Suchir Balaji, Shantanu Jain, William Saunders, Christopher Hesse, Andrew~N. Carr, Jan Leike, Josh Achiam, Vedant Misra, Evan Morikawa, Alec Radford, Matthew Knight, Miles Brundage, Mira Murati, Katie Mayer, Peter Welinder, Bob McGrew, Dario Amodei, Sam McCandlish, Ilya Sutskever, and Wojciech Zaremba.
\newblock Evaluating large language models trained on code, 2021.
\newblock URL \url{https://arxiv.org/abs/2107.03374}.

\bibitem[Cobbe et~al.(2021)Cobbe, Kosaraju, Bavarian, Chen, Jun, Kaiser, Plappert, Tworek, Hilton, Nakano, Hesse, and Schulman]{cobbe2021gsm8k}
Karl Cobbe, Vineet Kosaraju, Mohammad Bavarian, Mark Chen, Heewoo Jun, Lukasz Kaiser, Matthias Plappert, Jerry Tworek, Jacob Hilton, Reiichiro Nakano, Christopher Hesse, and John Schulman.
\newblock Training verifiers to solve math word problems.
\newblock \emph{arXiv preprint arXiv:2110.14168}, 2021.

\bibitem[Dalvi et~al.(2020)Dalvi, Sajjad, Durrani, and Belinkov]{dalvi-etal-2020-analyzing}
Fahim Dalvi, Hassan Sajjad, Nadir Durrani, and Yonatan Belinkov.
\newblock Analyzing redundancy in pretrained transformer models.
\newblock In Bonnie Webber, Trevor Cohn, Yulan He, and Yang Liu (eds.), \emph{Proceedings of the 2020 Conference on Empirical Methods in Natural Language Processing (EMNLP)}, pp.\  4908--4926, Online, November 2020. Association for Computational Linguistics.
\newblock \doi{10.18653/v1/2020.emnlp-main.398}.
\newblock URL \url{https://aclanthology.org/2020.emnlp-main.398/}.

\bibitem[Frantar \& Alistarh(2023)Frantar and Alistarh]{frantar2023sparsegpt}
Elias Frantar and Dan Alistarh.
\newblock Sparsegpt: Massive language models can be accurately pruned in one-shot.
\newblock In \emph{International Conference on Machine Learning (ICML)}, pp.\  10323--10337. PMLR, 2023.

\bibitem[Hu et~al.(2021)Hu, Shen, Wallis, Allen-Zhu, Li, Wang, Wang, and Chen]{hu2021loralowrankadaptationlarge}
Edward~J. Hu, Yelong Shen, Phillip Wallis, Zeyuan Allen-Zhu, Yuanzhi Li, Shean Wang, Lu~Wang, and Weizhu Chen.
\newblock Lora: Low-rank adaptation of large language models, 2021.
\newblock URL \url{https://arxiv.org/abs/2106.09685}.

\bibitem[Hui et~al.(2024)Hui, Yang, Cui, Yang, Liu, Zhang, Liu, Zhang, Yu, Lu, Dang, Fan, Zhang, Yang, Men, Huang, Zheng, Miao, Quan, Feng, Ren, Ren, Zhou, and Lin]{hui2024qwen25codertechnicalreport}
Binyuan Hui, Jian Yang, Zeyu Cui, Jiaxi Yang, Dayiheng Liu, Lei Zhang, Tianyu Liu, Jiajun Zhang, Bowen Yu, Keming Lu, Kai Dang, Yang Fan, Yichang Zhang, An~Yang, Rui Men, Fei Huang, Bo~Zheng, Yibo Miao, Shanghaoran Quan, Yunlong Feng, Xingzhang Ren, Xuancheng Ren, Jingren Zhou, and Junyang Lin.
\newblock Qwen2.5-coder technical report, 2024.
\newblock URL \url{https://arxiv.org/abs/2409.12186}.

\bibitem[Li et~al.(2025)Li, Li, Dong, and Liu]{li2025macromicroprobingdataset}
Haoyu Li, Xuhong Li, Yiming Dong, and Kun Liu.
\newblock From macro to micro: Probing dataset diversity in language model fine-tuning, 2025.
\newblock URL \url{https://arxiv.org/abs/2505.24768}.

\bibitem[Ma et~al.(2023)Ma, Fang, and Wang]{ma2023llmpruner}
Xinyin Ma, Gongfan Fang, and Xinchao Wang.
\newblock Llm-pruner: On the structural pruning of large language models.
\newblock In \emph{Advances in Neural Information Processing Systems (NeurIPS)}, volume~36, 2023.

\bibitem[{Multimodal Art Projection (M-A-P)}(2024)]{codefeedback_filtered_instruction_dataset}
{Multimodal Art Projection (M-A-P)}.
\newblock Codefeedback-filtered-instruction dataset.
\newblock \url{https://huggingface.co/datasets/m-a-p/CodeFeedback-Filtered-Instruction}, 2024.

\bibitem[{NVIDIA Corporation}(2023)]{nvidia2023matmult}
{NVIDIA Corporation}.
\newblock \emph{Matrix Multiplication Background User's Guide: Deep Learning Performance Documentation}, 2023.
\newblock URL \url{https://docs.nvidia.com/deeplearning/performance/dl-performance-matrix-multiplication/index.html}.
\newblock Accessed: 2026-02-19.

\bibitem[Pires et~al.(2023)Pires, Vilarinho~Lopes, Assogba, and Setiawan]{pires-etal-2023-one}
Telmo Pires, Ant{\'o}nio Vilarinho~Lopes, Yannick Assogba, and Hendra Setiawan.
\newblock One wide feedforward is all you need.
\newblock In Philipp Koehn, Barry Haddow, Tom Kocmi, and Christof Monz (eds.), \emph{Proceedings of the Eighth Conference on Machine Translation}, pp.\  1031--1044, Singapore, December 2023. Association for Computational Linguistics.
\newblock \doi{10.18653/v1/2023.wmt-1.98}.
\newblock URL \url{https://aclanthology.org/2023.wmt-1.98/}.

\bibitem[Pochinkov \& Schoots(2024)Pochinkov and Schoots]{pochinkov2024dissecting}
Nicholas Pochinkov and Nandi Schoots.
\newblock Dissecting language models: Machine unlearning via selective pruning.
\newblock \emph{arXiv preprint arXiv:2403.01267}, 2024.

\bibitem[Rajpurkar et~al.(2016)Rajpurkar, Zhang, Lopyrev, and Liang]{rajpurkar-etal-2016-squad}
Pranav Rajpurkar, Jian Zhang, Konstantin Lopyrev, and Percy Liang.
\newblock {SQ}u{AD}: 100,000+ questions for machine comprehension of text.
\newblock In Jian Su, Kevin Duh, and Xavier Carreras (eds.), \emph{Proceedings of the 2016 Conference on Empirical Methods in Natural Language Processing}, pp.\  2383--2392, Austin, Texas, November 2016. Association for Computational Linguistics.
\newblock \doi{10.18653/v1/D16-1264}.
\newblock URL \url{https://aclanthology.org/D16-1264}.

\bibitem[Reimers \& Gurevych(2019)Reimers and Gurevych]{reimers-gurevych-2019-sentence}
Nils Reimers and Iryna Gurevych.
\newblock Sentence-{BERT}: Sentence embeddings using {S}iamese {BERT}-networks.
\newblock In Kentaro Inui, Jing Jiang, Vincent Ng, and Xiaojun Wan (eds.), \emph{Proceedings of the 2019 Conference on Empirical Methods in Natural Language Processing and the 9th International Joint Conference on Natural Language Processing (EMNLP-IJCNLP)}, pp.\  3982--3992, Hong Kong, China, November 2019. Association for Computational Linguistics.
\newblock \doi{10.18653/v1/D19-1410}.
\newblock URL \url{https://aclanthology.org/D19-1410/}.

\bibitem[Shao et~al.(2024)Shao, Wang, Zhu, Xu, Song, Bi, Zhang, Zhang, Li, Wu, and Guo]{shao2024deepseekmathpushinglimitsmathematical}
Zhihong Shao, Peiyi Wang, Qihao Zhu, Runxin Xu, Junxiao Song, Xiao Bi, Haowei Zhang, Mingchuan Zhang, Y.~K. Li, Y.~Wu, and Daya Guo.
\newblock Deepseekmath: Pushing the limits of mathematical reasoning in open language models, 2024.
\newblock URL \url{https://arxiv.org/abs/2402.03300}.

\bibitem[Song et~al.(2024)Song, Zheng, Pan, Xue, Wang, Yin, Wang, Zhang, Chen, Chen, et~al.]{song2024powerinfer}
Yuzhong Song, Zeyu Zheng, Zhiying Pan, Chenhao Xue, Qipeng Wang, Jianwei Yin, Rui Wang, Haotian Zhang, Wenju Chen, Mingsong Chen, et~al.
\newblock Powerinfer: Fast large language model inference with macroscopic sparsity.
\newblock In \emph{Proceedings of the 29th ACM SIGPLAN Annual Symposium on Principles and Practice of Parallel Programming (PPoPP)}, pp.\  411--425, 2024.
\newblock URL \url{https://arxiv.org/abs/2312.12456}.

\bibitem[Suriya(2024)]{everydayConversationalCleaned2024}
Suriya.
\newblock everyday-conversational-cleaned.
\newblock Dataset hosted on Hugging Face, 2024.
\newblock URL \url{https://huggingface.co/datasets/suriya7/everyday-Conversational-cleaned}.

\bibitem[Wang et~al.(2024)]{wang2024sharing}
Weixuan Wang et~al.
\newblock Sharing matters: Analysing neurons across languages and tasks in {LLMs}.
\newblock \emph{arXiv preprint arXiv:2406.09265}, 2024.

\bibitem[Xia et~al.(2024)Xia, Gao, Zeng, and Chen]{xia2024sheared}
Mengzhou Xia, Tianyu Gao, Zhiyuan Zeng, and Danqi Chen.
\newblock Sheared llama: Accelerating language model pre-training via structured pruning.
\newblock In \emph{The Twelfth International Conference on Learning Representations (ICLR)}, 2024.

\bibitem[Yang et~al.(2024)Yang, Zhang, Hui, Gao, Yu, Li, Liu, Tu, Zhou, Lin, Lu, Xue, Lin, Liu, Ren, and Zhang]{yang2024qwen25mathtechnicalreportmathematical}
An~Yang, Beichen Zhang, Binyuan Hui, Bofei Gao, Bowen Yu, Chengpeng Li, Dayiheng Liu, Jianhong Tu, Jingren Zhou, Junyang Lin, Keming Lu, Mingfeng Xue, Runji Lin, Tianyu Liu, Xingzhang Ren, and Zhenru Zhang.
\newblock Qwen2.5-math technical report: Toward mathematical expert model via self-improvement.
\newblock \emph{arXiv preprint arXiv:2409.12122}, 2024.

\bibitem[Yu et~al.(2022)Yu, Khadivi, and Xu]{yu-etal-2022-data}
Yu~Yu, Shahram Khadivi, and Jia Xu.
\newblock Can data diversity enhance learning generalization?
\newblock In Nicoletta Calzolari, Chu-Ren Huang, Hansaem Kim, James Pustejovsky, Leo Wanner, Key-Sun Choi, Pum-Mo Ryu, Hsin-Hsi Chen, Lucia Donatelli, Heng Ji, Sadao Kurohashi, Patrizia Paggio, Nianwen Xue, Seokhwan Kim, Younggyun Hahm, Zhong He, Tony~Kyungil Lee, Enrico Santus, Francis Bond, and Seung-Hoon Na (eds.), \emph{Proceedings of the 29th International Conference on Computational Linguistics}, pp.\  4933--4945, Gyeongju, Republic of Korea, October 2022. International Committee on Computational Linguistics.
\newblock URL \url{https://aclanthology.org/2022.coling-1.437/}.

\bibitem[Zhang et~al.(2020)Zhang, Kishore, Wu, Weinberger, and Artzi]{zhang2020bertscoreevaluatingtextgeneration}
Tianyi Zhang, Varsha Kishore, Felix Wu, Kilian~Q. Weinberger, and Yoav Artzi.
\newblock Bertscore: Evaluating text generation with bert, 2020.
\newblock URL \url{https://arxiv.org/abs/1904.09675}.

\bibitem[Zhang et~al.(2024)]{zhang2024loraprune}
Yiming Zhang et~al.
\newblock Loraprune: Pruning meets low-rank parameter-efficient fine-tuning.
\newblock In \emph{The Twelfth International Conference on Learning Representations (ICLR)}, 2024.

\end{thebibliography}
\bibliographystyle{main}

\section*{Reproducibility Statement}
The datasets used in this work are described in~\nameref{sec:3.1}. To facilitate reproducibility, we provide complete experimental details in the supplementary material, including the hardware/software environment (Python 3.12.3, PyTorch 2.5.1+cu121, NVIDIA RTX A6000 48GB), LoRA fine-tuning hyperparameters for both Qwen-Code and Qwen-Math models (Table~\ref{tab:full_hyperparameters}), the full HumanEval evaluation configuration (Table~\ref{tab:humaneval_config}), and the configurations used for semantic similarity metrics including BERTScore and SBERT cosine similarity (Tables~\ref{tab:bertscore_config} and \ref{tab:sbert_config}). We also include the exact~\nameref{section:instruction} used in all experiments. Additionally, we provide an anonymous repository that contains all the codes/scripts used in our research and our supplementary data at: \url{https://anonymous.4open.science/r/task-specific-pruning-EC02/}.

\section*{AI Usage Statement}
We used AI-assisted tools to improve writing quality, including grammatical correction and language polishing. We also used GitHub Copilot to assist with code development. All experimental data were generated through our own experiments and pipelines. No results, claims, or empirical findings were generated by AI tools. All interpretations, analyses, and conclusions were produced by the authors, with AI tools used only to polish the presentation of the analysis.

\appendix
\renewcommand{\numberline}[1]{}
\section{Experimental Setup}
\begin{table}[h]
\centering
\begin{tabular}{ll}
\hline
\textbf{Component} & \textbf{Specification} \\
\hline
Programming Language & Python 3.12.3 \\
Framework & PyTorch 2.5.1+cu121 \\
GPU & NVIDIA RTX A6000 (48 GB VRAM) \\
CPU & Intel i9-14900K \\
System Memory & 64 GB RAM \\
\hline
\end{tabular}
\label{tab:hardware_setup}
\caption{Experimental Hardware and Software Setup}
\end{table}

\section{Extended Mathematical Formulation of Activation-Based Structural Pruning}
\label{app:math_formulation}

To ensure full reproducibility and provide a rigorous theoretical grounding for our methodology, this section details the extended mathematical formulations governing activation capture, neuron selectivity, and physical matrix reduction.

\subsection{Notation and Dimensionality}
Let $N$ denote the number of sequences (prompts), $L$ the number of transformer layers, and $d$ the model hidden size. The intermediate (feed-forward) width is $d_{ff}$, which equals the number of MLP neurons per layer. For a fixed layer $\ell \in \{0,\dots,L-1\}$, neurons are indexed by $n \in \{1,\dots,N_\ell\}$ with $N_\ell = d_{ff}$. Unless stated otherwise, all activations are computed in inference mode.

\subsection{Masked Activation Aggregation}
The SwiGLU intermediate activation $h^{(\ell)}$ captures the interaction between the gate and up-projected input and is central to understanding the neuron's contribution to the MLP output. We take the element-wise absolute value $a^{(\ell)} = \lvert h^{(\ell)} \rvert$. Since sequences may include padded tokens, we compute a masked mean activation across valid (non-padded) tokens. For the $i$-th input sequence with $T_i$ tokens, the mean activation is:
\begin{equation}
    \bar{a}^{(\ell)}_i = \frac{\sum_{t=1}^{T_i} a^{(\ell)}_{i,t} \cdot m_{i,t}}{\sum_{t=1}^{T_i} m_{i,t}},
\end{equation}
where $a^{(\ell)}_{i,t} \in \mathbb{R}^{d_{ff}}$ is the absolute activation vector for token $t$ in sequence $i$ at layer $\ell$, and $m_{i,t} \in \{0,1\}$ is the attention mask. The denominator is a scalar and divides the vector numerator component-wise (broadcasting), yielding $\bar{a}^{(\ell)}_i \in \mathbb{R}^{d_{ff}}$. While per-token activation would offer more depth, we employ prompt-level averaging as a practical compromise to retain core activation trends while significantly reducing memory and processing demands under our hardware constraints. We construct a dataset-wide tensor of mean activations:
\begin{equation}
    A \in \mathbb{R}^{N \times L \times d_{ff}}, \quad \text{with} \quad A[i, \ell, :] = \bar{a}^{(\ell)}_i.
\end{equation}

\subsection{Formal Thresholding and Dimensional Reduction}
For each neuron, we calculate a selectivity score defined as the standardized difference in its average activations between target and distractor prompts. A score where $S^{(\ell)}_n > 0$ means the neuron activates more for target prompts, while $S^{(\ell)}_n < 0$ means it activates more for distractor prompts (candidate for pruning). 

To determine the precise ablation set, we define a layer-specific threshold $\theta^{(\ell)}$. Structural pruning is applied uniformly across all MLP layers such that the remaining neuron counts in each layer remain divisible by 128 to improve matrix multiplication efficiency in GPU kernels. The threshold $\theta^{(\ell)}$ is strictly calibrated to satisfy this hardware divisibility constraint. The set of neurons to prune is:
\begin{equation}
    \mathcal{P}^{(\ell)} = \{\, n \in \{1,\dots,N_\ell\} \mid S^{(\ell)}_n \le \theta^{(\ell)} \,\}.
\end{equation}

The surviving set is the disjoint complement:
\begin{equation}
    \mathcal{K}^{(\ell)} = \{1,\dots,N_\ell\} \setminus \mathcal{P}^{(\ell)}.
\end{equation}

Pruning physically removes neurons in $\mathcal{P}^{(\ell)}$ from the MLP weight matrices. In the row-based mathematical convention, each neuron corresponds to one row in $W^{(\ell)}_{\text{gate}}$ and $W^{(\ell)}_{\text{up}}$, and one column in $W^{(\ell)}_{\text{down}}$. We apply the same survivor index set $\mathcal{K}^{(\ell)}$ consistently across all three matrices, preserving the original neuron order:
\begin{align}
    \tilde{W}^{(\ell)}_{\text{gate}} &= W^{(\ell)}_{\text{gate}}[\mathcal{K}^{(\ell)}, :] \in \mathbb{R}^{\lvert \mathcal{K}^{(\ell)} \rvert \times d}, \\
    \tilde{W}^{(\ell)}_{\text{up}}   &= W^{(\ell)}_{\text{up}}[\mathcal{K}^{(\ell)}, :] \in \mathbb{R}^{\lvert \mathcal{K}^{(\ell)} \rvert \times d}, \\
    \tilde{W}^{(\ell)}_{\text{down}} &= W^{(\ell)}_{\text{down}}[:, \mathcal{K}^{(\ell)}] \in \mathbb{R}^{d \times \lvert \mathcal{K}^{(\ell)} \rvert}.
\end{align}

We keep the rows/columns indexed by $\mathcal{K}^{(\ell)}$, thereby reducing the intermediate dimension of the MLP from $N_\ell$ to $\lvert \mathcal{K}^{(\ell)} \rvert$.

\section{Extended Fine-Tuning Configurations}
\label{app:hyperparameters}

In~\nameref{sec:3.5}, we outlined the core Low-Rank Adaptation (LoRA) hyperparameters utilized to recover task-specific performance after structural pruning. Table~\ref{tab:full_hyperparameters} provides the complete, exhaustive enumeration of the training configurations applied to both the Code and Math models to ensure full reproducibility.\\

\begin{table}[h]
    \centering
    \begin{tabular}{@{}lcc@{}}
        \toprule
        \textbf{Hyperparameter} & \textbf{Code Models (Qwen-Code)} & \textbf{Math Models (Qwen-Math)} \\
        \midrule
        Precision & Bfloat16 & Bfloat16 \\
        Epochs & 2 & 2 \\
        Learning Rate & $2 \times 10^{-4}$ & $2 \times 10^{-5}$ \\
        LoRA Rank ($r$) & 16 & 8 \\
        LoRA Alpha ($\alpha$) & 32 & 16 \\
        LoRA Dropout & 0.10 & 0.10 \\
        Effective Batch Size & 16 & 8 \\
        Target Modules & \texttt{q\_proj, k\_proj, v\_proj,} & \texttt{q\_proj, k\_proj, v\_proj,} \\
                       & \texttt{o\_proj, gate\_proj,} & \texttt{o\_proj, gate\_proj,} \\
                       & \texttt{up\_proj, down\_proj} & \texttt{up\_proj, down\_proj} \\
        \bottomrule
    \end{tabular}
    \caption{Complete LoRA Fine-Tuning Hyperparameters}
    \label{tab:full_hyperparameters}
\end{table}

\section{HumanEval Configuration}

Since the model often produces natural language explanations in addition to the code, we strip all non-code content and retain only the generated code block when evaluating functional correctness.

\vspace{20pt}
\begin{table}[H]
\centering
\begin{tabular}{ll}
\hline
\textbf{Parameter} & \textbf{Setting} \\
\hline
Decoding Strategy & Greedy (Pass@1) \\
Max New Tokens & 512 \\
Sampling & Disabled (do\_sample=False) \\
Beam Search & 1 (num\_beams=1) \\
Stopping Criterion & EOS token \\
Batch Size & 16 \\
Timeout per Task & 3.0 seconds \\
Number of Tasks & 164 \\
Dataset Split & Test (official prompts) \\
Evaluation Metric & Functional Correctness (Pass@1) \\
Sanitization & Remove markdown \& explanatory text(keep only code block) \\
\hline
\end{tabular}
\caption{HumanEval Evaluation Configuration}
\label{tab:humaneval_config}
\end{table}

\section[model prompting instructions]{Model Instruction Format}
\label{section:instruction}
This section documents the exact prompting instructions used for different evaluation settings.

\subsection{Mathematical Reasoning Instruction}

\noindent\textbf{Instruction:}
\begin{quote}
Please reason step by step, and put your final answer within \textbackslash boxed\{\}.
Then we use regex to retrive the actual answer.
\end{quote}

\subsection{Code Generation Instruction (HumanEval)}
\noindent\textbf{Instruction:}
\begin{quote}
You are generating solutions for the HumanEval benchmark. Output ONLY valid Python code. Output ONLY the function definition requested. Do NOT include explanations, comments outside the function, examples, tests, prints, or prose. Do NOT include markdown, backticks, or formatting. Do NOT add helper text before or after the code. The output must consist of one or more top-level def functions only. No top-level statements other than function definitions. No if \_\_name\_\_ == ``\_\_main\_\_'' blocks. No print statements. No test cases. No explanations. Respond with Python code ONLY.
\end{quote}

\noindent After generation, any remaining text or markdown is removed by a custom sanitization script, retaining only the code for HumanEval evaluation, as the benchmark accepts only function definitions and additional text leads to failure.

\section[Examples of Traps]{Examples of Traps}
\label{sec:trap_examples}

In this section, we provide representative examples of degeneration traps observed under unconstrained generation. These examples illustrate the two trap types described in the main text, along with a mitigated case for Type~1.

\subsection{Type 1: Degeneration Loop with Complete Reasoning}

\begin{figure}[H]
    \centering
    \captionsetup{font=small}
    
    \begin{minipage}{0.48\linewidth}
        \centering
        \includegraphics[width=\linewidth]{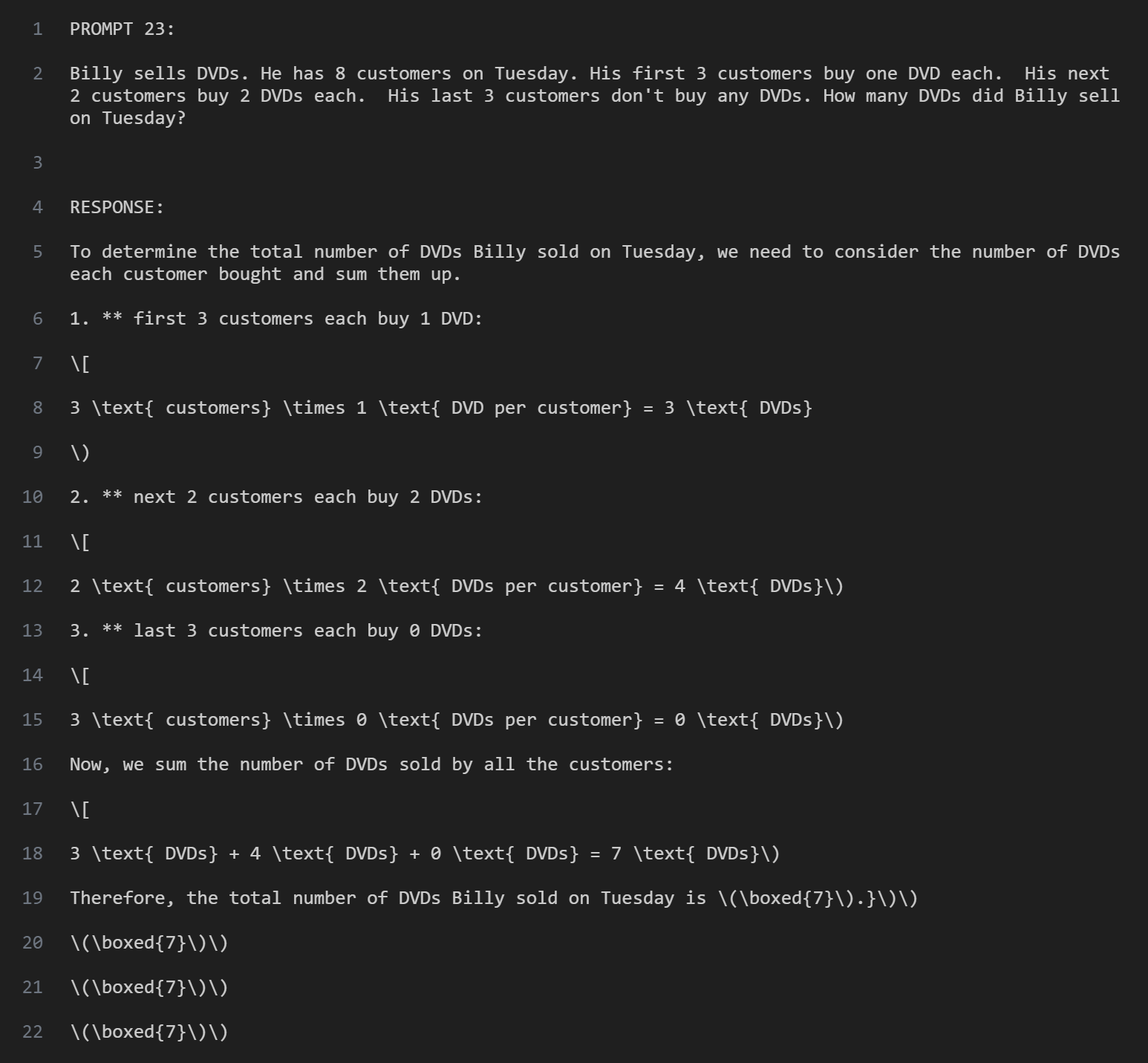}
    \end{minipage}
    \hfill
    \begin{minipage}{0.48\linewidth}
        \centering
        \includegraphics[width=\linewidth]{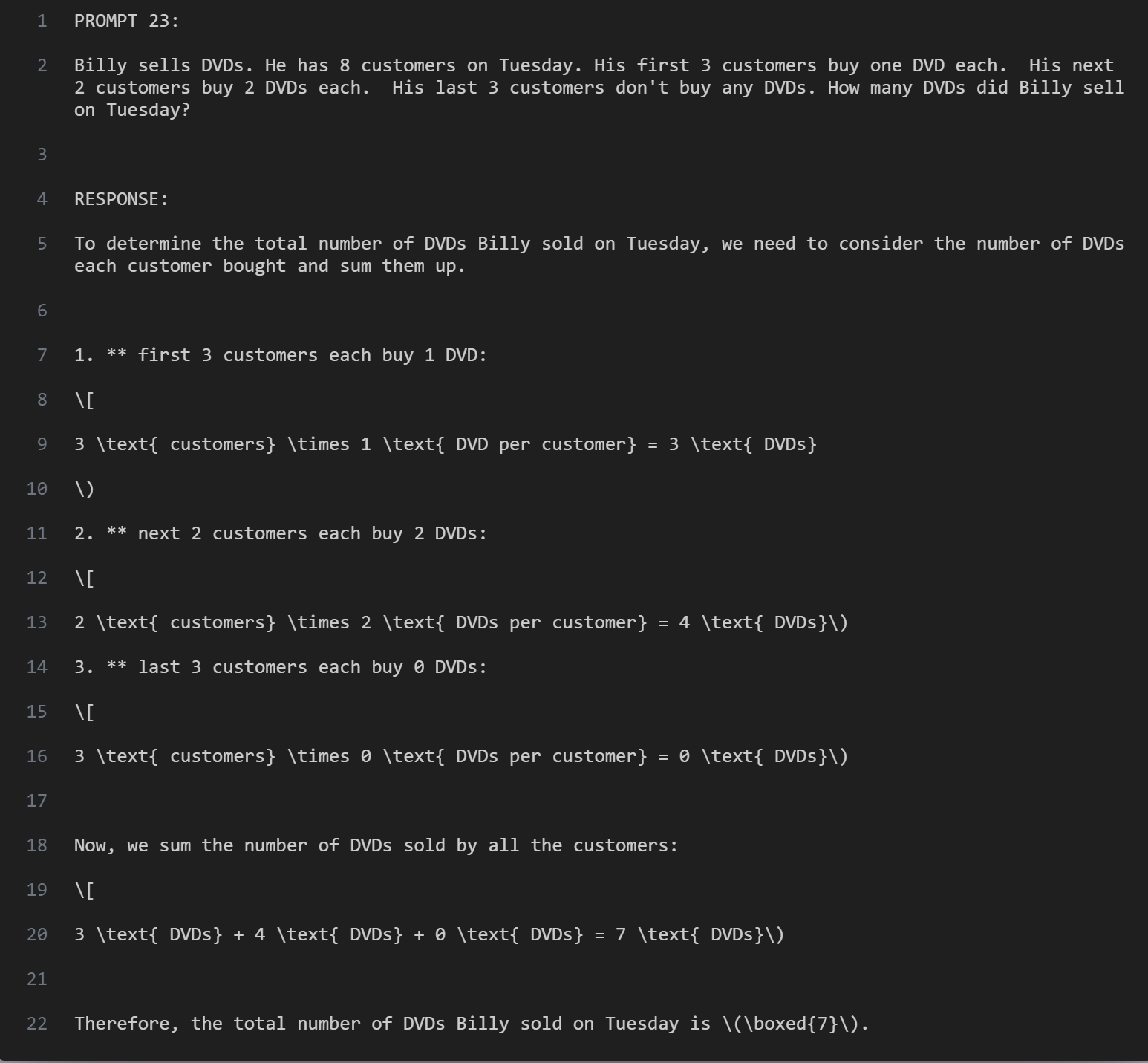}
    \end{minipage}
    
    \caption{Example of a Type~1 trap (left) and mitigated Type~1 case (right).}
    \label{fig:type1_trap}
\end{figure}

In Type~1 traps, the model produces meaningful reasoning and a correctly formatted final answer, but fails to emit an end-of-sequence (EOS) token. Instead, generation continues with repeated symbols, tokens, or sequences.

Figure~\ref{fig:type1_trap} shows an example where the model completes the reasoning process and outputs the correct answer, followed by a degeneration loop.

\subsection{Mitigated Type 1 Example}

Type~1 traps can be mitigated in practice by using custom stopping criteria. With our custom criterion, the model terminates immediately after producing the final answer, preventing degeneration loops. Specifically, generation stops once the output contains a complete LaTeX boxed expression, i.e., a substring of the form: \begin{verbatim} \(\boxed{...}\) \end{verbatim}

Figure~\ref{fig:type1_trap} (right) shows a mitigated example of a Type~1 case where generation is properly terminated using the custom stopping criteria.

\vspace{32pt}
\subsection{Type 2: Degeneration Loop with Incomplete Reasoning}
\setlength{\intextsep}{4pt}

\begin{wrapfigure}[16]{r}{0.4\columnwidth} 
\vspace{-8pt}
\centering
\includegraphics[width=\linewidth]{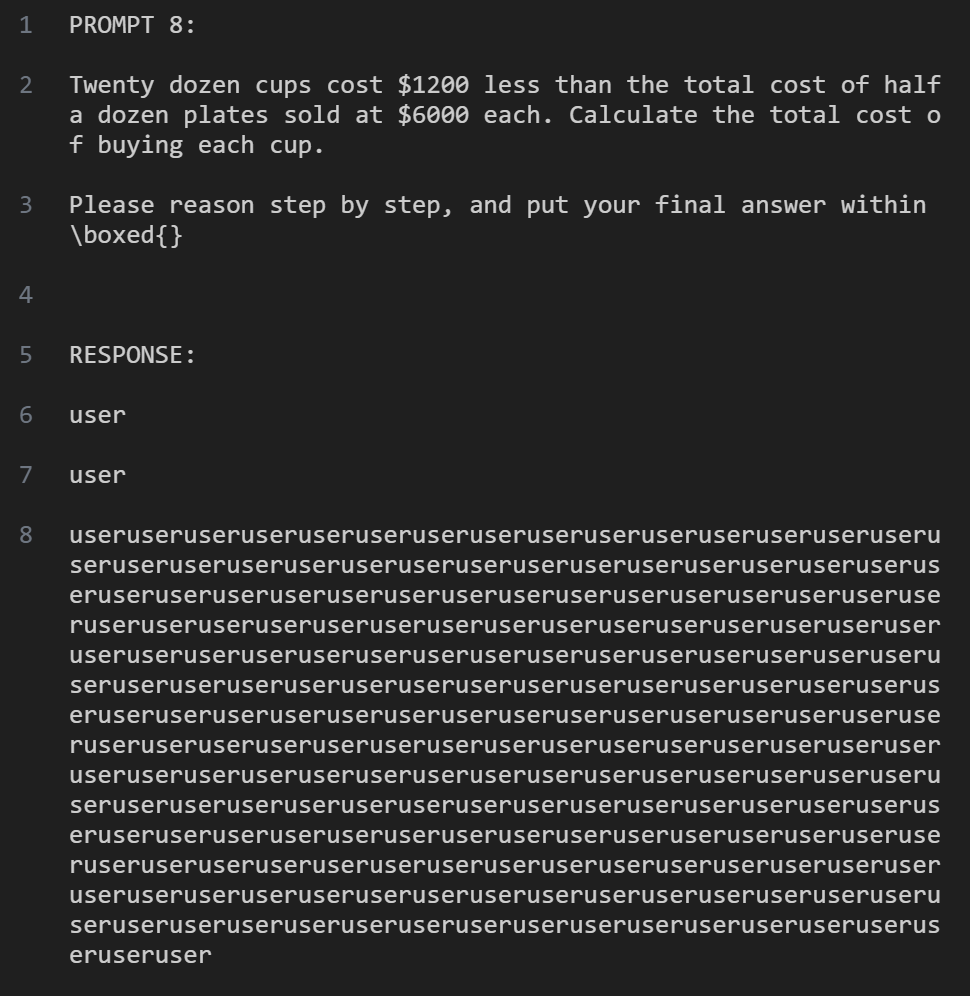}
\vspace{-6pt} 
\captionsetup{font=scriptsize}
\caption{Example of a Type~2 trap.}
\label{fig:type2_trap}
\vspace{-5pt}
\end{wrapfigure}

In Type~2 traps, the model fails to produce meaningful reasoning. The generation collapses early or mid-sequence into incoherent and nonsensical text, often characterized by repeated symbols or tokens.

Figure~\ref{fig:type2_trap} presents an example where the model fails to complete reasoning and enters a degeneration loop prematurely.

\subsection{Discussion}

These examples highlight that degeneration loops arise from failures in EOS prediction and reasoning stability. Type~1 traps primarily reflect termination failure after successful reasoning, while Type~2 traps indicate disruption of the reasoning process itself. As observed in the main text, increasing pruning ratio leads to a higher frequency of such behaviors, particularly Type~2 traps.

\section{BERTScore and SBERT Cosine Similarity Configuration}

Instead of using other metrics for distractor task, we evaluate performance using semantic similarity, as our goal is to quantify how much the pruned model’s responses deviate from the original model. Greater deviation from the original responses is reflected in the similarity score. Specifically, we use BERTScore for QnA tasks and SBERT cosine similarity for dialogue tasks.

\vspace{20pt}
\begin{table}[H]
\centering
\small
\begin{tabular}{lc}
\hline
\textbf{Configuration} & \textbf{BERTScore} \\
\hline
Model & Pretrained BERT \\
Rescale with Baseline & Yes (internal adjustment) \\
Benchmark / Task & QnA task \\
\hline
\end{tabular}
\caption{BERTScore evaluation configuration}
\label{tab:bertscore_config}
\end{table}

\begin{table}[H]
\vspace{6pt}
\centering
\small
\begin{tabular}{lc}
\hline
\textbf{Configuration} & \textbf{SBERT Cosine Similarity} \\
\hline
Model & all-mpnet-base-v2 \\
Rescale with Baseline & N/A \\
Benchmark / Task & Dialogue task \\
\hline
\end{tabular}
\caption{SBERT evaluation configuration}
\label{tab:sbert_config}
\end{table}

\end{document}